\documentclass[letterpaper]{article} 
\usepackage[draft]{aaai2026}  
\usepackage{times}  
\usepackage{helvet}  
\usepackage{courier}  
\usepackage[hyphens]{url}  
\usepackage{graphicx} 
\usepackage{natbib}  
\usepackage{caption} 
\usepackage{algorithm}
\usepackage{algorithmic}

\usepackage{newfloat}
\usepackage{listings}
\DeclareCaptionStyle{ruled}{labelfont=normalfont,labelsep=colon,strut=off} 
\floatstyle{ruled}
\newfloat{listing}{tb}{lst}{}
\floatname{listing}{Listing}
\usepackage{MyPreamble}

\newcommand\TODO[1][]{{\color{orange}[TODO\ifthenelse{\equal{#1}{}}{}{: #1}]}}
\newcommand\Ours{GRAPHITE}
\newcommand\OursFull{\emph{\underline{GRA}ph homo\underline{PHI}ly boos\underline{TE}r}}
\newcommand\EQref[1]{Equation~\eqref{#1}}

\title{Graph Homophily Booster: Rethinking the Role of\\Discrete Features on Heterophilic Graphs}
\author{
    Ruizhong Qiu\textsuperscript{\rm 1}\equalcontrib,
    Ting-Wei Li\textsuperscript{\rm 1}\equalcontrib,
    Gaotang Li\textsuperscript{\rm 1},
    Hanghang Tong\textsuperscript{\rm 1} 
}
\affiliations{
    \textsuperscript{\rm 1}University of Illinois Urbana--Champaign\\
    \{rq5, twli, gaotang3, htong\}@illinois.edu

}

\usepackage{bibentry}

\begin{document}

\maketitle


\begin{abstract}

Graph neural networks (GNNs) have emerged as a powerful tool for modeling graph-structured data, demonstrating remarkable success in many real-world applications such as complex biological network analysis, neuroscientific analysis, and social network analysis. However, existing GNNs often struggle with heterophilic graphs, where connected nodes tend to have dissimilar features or labels. While numerous methods have been proposed to address this challenge, they primarily focus on architectural designs without directly targeting the root cause of the heterophily problem. These approaches still perform even worse than the simplest MLPs on challenging heterophilic datasets. For instance, our experiments show that 21 latest GNNs still fall behind the MLP on the \textsc{Actor} dataset. This critical challenge calls for an innovative approach to addressing graph heterophily beyond architectural designs. To bridge this gap, we propose and study a new and unexplored paradigm: \emph{directly} increasing the graph homophily via a carefully designed graph transformation. In this work, we present a simple yet effective framework called \OursFull{} (\Ours{}) to address graph heterophily. To the best of our knowledge, this work is the first method that explicitly transforms the graph to directly improve the graph homophily. Stemmed from the exact definition of homophily, our proposed \Ours{} creates \emph{feature nodes} to facilitate homophilic message passing between nodes that share similar features.  Furthermore, we both theoretically and empirically show that our proposed \Ours{} significantly increases the homophily of originally heterophilic graphs, with only a slight increase in the graph size. Extensive experiments on challenging datasets demonstrate that our proposed \Ours{} significantly outperforms state-of-the-art methods on heterophilic graphs while achieving comparable accuracy with state-of-the-art methods on homophilic graphs. Furthermore, our proposed graph transformation alone can already enhance the performance of homophilic GNNs on heterophilic graphs, even though they were not originally designed for heterophilic graphs. 
We will release our code upon the publication of this paper. 
\end{abstract}

\section{Introduction}



Graph neural networks (GNNs) have emerged as a powerful class of models for learning 
on topologically structured data. Their ability to incorporate both graph topology and node-level attributes has enabled them to achieve state-of-the-art results in a wide range of applications. These include protein function prediction, where GNNs model complex biological networks~\cite{you2021deepgraphgo,reau2023deeprank}; neuroscientific analysis, where they are used to model brain networks~\cite{li2023interpretable}; and social network analysis, where they help uncover patterns and relationships among users~\cite{li2023survey}.

A critical challenge that many GNNs are faced with is that real-world networks can exhibit heterophily, where connected nodes tend to have dissimilar featuresor labels. Examples include protein--protein interaction networks where different types of proteins interact~\cite{zhu2020beyond}, or online marketplace networks where buyers connect with sellers rather than other buyers~\cite{pandit2007netprobe}. Standard GNN architectures~\cite{kipf2016semi, wu2019simplifying, velivckovic2017graph, hamilton2017inductive, chen2020simple, abu2019mixhop}, with their heavy reliance on neighborhood aggregation, often struggle with heterophilous graphs since aggregating features from dissimilar neighbors can dilute or distort node representations.
Existing methods for heterophilic graphs mainly focus on designing new GNN architectures as workarounds for heterophilic graphs, such as separating ego and neighbor embeddings \cite{zhu2020beyond}, incorporating higher-order information with learnable weights \cite{chien2020adaptive}, and adaptive self-gating to leverage both low- and high-frequency signals \cite{bo2021beyond}. More recent solutions introduce frequency-based filtering to handle both homophily and heterophily or leverage adaptive residual connections to further enhance flexibility \cite{xu2023node, xuslog, yan2024trainable}.

Despite plenty of architectural advances, many GNNs still perform even worse than the simplest multi-layer perceptrons (MLPs) on challenging heterophilic graphs. For instance, Table~\ref{tab:main-res} shows that 21 latest GNNs still fall behind the MLP on the \textsc{Actor} dataset. This critical challenge calls for an innovative approach to addressing graph heterophily beyond architectural designs. 

To bridge this gap, we propose and study a new and unexplored paradigm: \emph{directly} increasing the graph homophily via a carefully designed graph transformation. In this work, we present a simple yet effective framework called \OursFull{} (\Ours{}) to address graph heterophily. To the best of our knowledge, this work is the first method that explicitly transforms the graph to directly improve the graph homophily. 

\begin{figure}[t]
\hfill
\includegraphics[width=\linewidth]{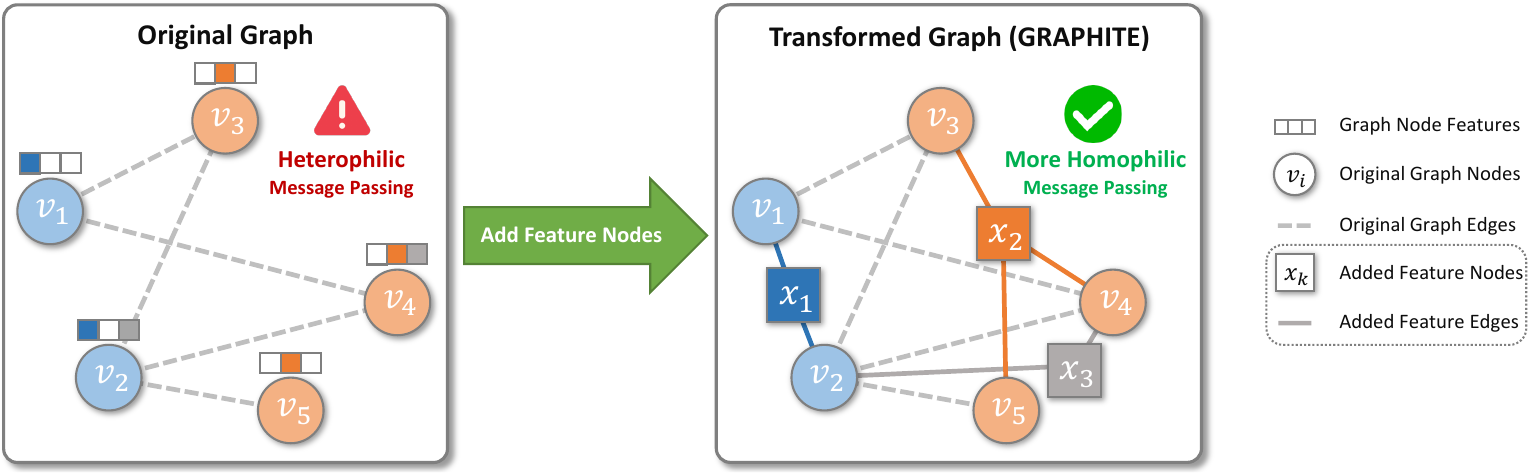}
\caption{Overview of our proposed \Ours{}. The added feature nodes can facilitate homophilic message passing. For instance, feature node $x_1$ facilitates homophilic message passing between graph nodes $v_1,v_2$, and feature node $x_2$ facilitates homophilic message passing among graph nodes $v_3,v_4,v_5$. 
}
\label{fig:illus}
\vspace{-1em}
\end{figure}

Our key idea is rooted in the exact definition of homophily and heterophily. In a homophilic/heterophilic graph, nodes that share similar features are more/less likely to be adjacent, respectively. Therefore, a natural idea to increase the graph homophily is to create ``shortcut" connections between nodes with similar features so as to facilitate homophilic message passing between them. However, na\"ively adding mutual connections between such node pairs can drastically increase the number of edges. For example, even if a graph has only 2,000 nodes, the na\"ive approach can add as many as 1,999,000 ``shortcut'' edges. To reduce the number of ``shortcut'' edges, we propose to connect such node pairs \emph{indirectly} instead. In particular, we introduce \emph{feature nodes} as ``hubs'' and connect graph nodes to their corresponding feature nodes. We further theoretically show that our proposed method can provably enhance the homophily of originally heterophilic graphs without increasing the graph size much. 

Our main contributions are summarized as follows:
\begin{itemize}
    \item\textbf{New paradigm.} We propose and study a new and unexplored paradigm: \emph{directly} increasing the graph homophily via graph transformation. This paper is the first work on this paradigm to the best of our knowledge. 
    \item\textbf{Proposed method.} We propose a simple yet effective method called \Ours{}, which creates feature nodes as ``shortcuts'' to facilitate homophilic message passing between nodes with similar features. 
    \item\textbf{Theoretical guarantees.} We theoretically show that our proposed \Ours{} can \emph{provably} enhance the homophily of originally heterophilic graphs with only a \emph{slight} increase in the graph size. 
    \item\textbf{Empirical performance.} Extensive experiments on challenging datasets demonstrate the effectiveness of our proposed \Ours{}. Our proposed \Ours{} \emph{significantly} outperforms state-of-the-art methods on heterophilic graphs while achieving \emph{comparable} accuracy with state-of-the-art methods on homophilic graphs. Furthermore, our proposed graph transformation alone can already enhance the performance of homophilic GNNs on heterophilic graphs. 
\end{itemize}

\section{Preliminaries}
\label{sec:prob-def}


\subsection{Notation}


An undirected graph with discrete node features can be represented as a triple $\mathcal{G} = ( \mathcal{V}, \mathcal{E}, \BM X )$, where $\mathcal{V}=\{v_1,\dots,v_{|\CAL V|}\}$ denotes the node set, $\mathcal{E}\subseteq\CAL V\times\CAL V$ denotes the edge set, $\BM X \in \{0,1\}^{\mathcal{V}\times \mathcal{X}}$ is a binary node feature matrix representing discrete node features, and $\mathcal{X}=\{1,\dots,|\CAL X|\}$ is the feature set containing all the discrete node features. In addition to that, each graph node $v_i \in \mathcal{V}$ has a node label $y_{v_i} \in\mathcal{Y}$, where $\mathcal{Y}$ is the label set with $C = |\mathcal{Y}|$ classes.




\subsection{Problem Definition}

In this paper, we study two key problems: (i) how to transform a graph to increase its homophily and (ii) how to perform node classification on a heterophilic graph datasets. Formally, we introduce the problem definitions as follows.

\begin{PRB}[Boosting Graph Homophily] Given a highly heterophilic graph, transform the graph to increase its homophily. \textbf{Input:} a heterophilic graph $\mathcal{G}$. \textbf{Output:} a transformed graph $\mathcal{G}^*$  with higher homophily.
\label{def:homo-boost}
\end{PRB}


\begin{PRB}[Semi-supervised Node Classification on a Heterophilic Graph]
Given a heterophilic graph and a set of labelled nodes, train a model to predict the labels of unlabelled nodes. \textbf{Input:} (i) a heterophilic graph \( \mathcal{G}=(\CAL V,\CAL E,\BM X)\); (ii) a labelled node set \( \mathcal{V}_\textnormal{L} \subset \mathcal{V} \) whose node labels $[y_{v_i}]_{v_i\in\CAL V_\textnormal L}$ are available. \textbf{Output:} the predicted labels of unlabeled nodes \( \mathcal{V} \setminus \mathcal{V}_\textnormal{L} \).

\label{def:node-classify}
\end{PRB}



\begin{figure*}[t]
\centering
\includegraphics[width=0.64\linewidth]{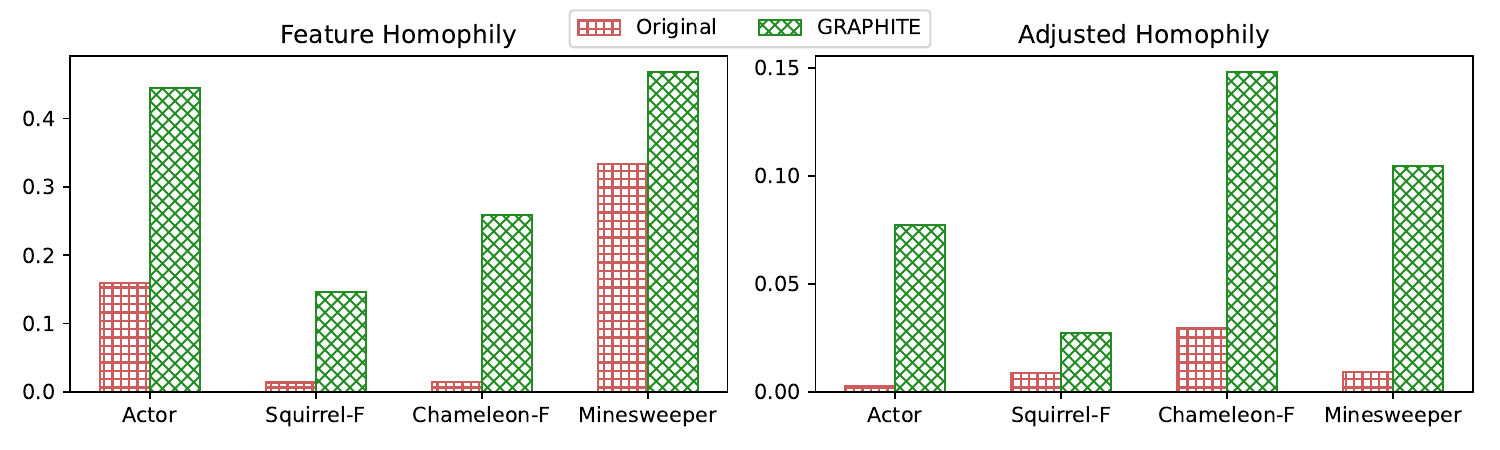}
\caption{Our proposed \Ours{} significantly increases the homophily of originally heterophilic graphs. We report two latest homophily metrics: \emph{feature homophily} \cite{jin2022raw} and \emph{adjusted homophily} \cite{platonov2024characterizing}. 
}
\label{fig:homo-analysis}
\vspace{-1em}
\end{figure*}

\section{Proposed Method: \Ours{}}
In this section, we propose a simple yet effective graph transformation method called \OursFull{} (\Ours{}) that can efficiently increase the homophily of a graph. In Section~\ref{ssec:motiv}, we will introduce the motivation of our proposed \Ours{}. 
First, we will present the design of our proposed method \Ours{}. 
Then, we will describe the neural architecture of our proposed method. Due to the page limit, proofs of theoretical results are deferred to the appendix.

\subsection{Motivation}\label{ssec:motiv}

Graph heterophily is a ubiquitous challenge in graph-based machine learning. On a highly heterophilic graph, many neighboring nodes exhibit dissimilar features or belong to different classes. As a result, graph heterophily limits the effectiveness of GNN message passing, as standard aggregation schemes might fail to capture meaningful patterns in heterophilic neighbors.

Existing methods for heterophilic graphs mainly focus on designing workarounds such as new architectures or learning paradigms for heterophilic graphs, including adaptive message passing, higher-order neighborhoods, or alternative propagation mechanisms that leverage both local and global graph structures. 

In contrast to existing workaround methods, we propose a new method that aims to directly increase the homophily of the graph via a specially designed graph transformation. To the best of our knowledge, this work is the first method that explicitly transforms the graph to improve the homophily of the graph. 

Our idea is rooted in the exact definition of homophily and heterophily. In a heterophilic graph, nodes that share similar features are more likely to be non-adjacent. However, in a homophilic graph, nodes that share similar features should be more likely to be neighbors. Therefore, a natural idea to increase the homophily of the graph is to create ``shortcut" connections between nodes with similar features, which will facilitate homophilic message passing between them.

Before we introduce the proposed method, let's consider the following na\"ive approach to implementing the aforementioned idea: For each pair of nodes $v_i,v_j \in\CAL V$, if they share at least a feature (i.e., $\|\BM X[v_i,:]\land\BM X[v_j,:]\|_\infty>0$), we add a ``shortcut'' edge $(v_i,v_j)$ between them. Let's call this approach the \emph{na\"ive homophily booster} (NHB). The following Theorem~\ref{thm:naive} shows that NHB can indeed increase the homophily of the graph under mild and realistic assumptions. 

\begin{THM}[Na\"ive homophily booster]\label{thm:naive}
Given a heterophilic graph $\CAL G=(\CAL V,\CAL E,\BM X)$, let $\CAL E^\dagger$ denote the set of edges after adding the NHB ``shortcut'' edges, and let $\CAL G^\dagger:=(\CAL V,\CAL E^\dagger,\BM X)$ denote the graph transformed by NHB. Under mild and realistic assumptions in Appendix~\ref{app:ass}, we have
\GA{
\OP{hom}(\CAL G^\dagger)>\OP{hom}(\CAL G),\\
|\CAL E^\dagger|-|\CAL E|\le O(|\CAL V|^2).\label{eq:naive-edges}
}
\end{THM}

However, \EQref{eq:naive-edges} also shows that NHB is extremely inefficient despite its effectiveness in increasing homophily. For instance, even if the graph has only 2,000 nodes, NHB can add as many as 1,999,000 ``shortcut'' edges. The plenty of ``shortcut'' edges can drastically slow down the training and the inference process of GNNs. Hence, this na\"ive approach is computationally impractical for GNNs. To address this computational challenge, we will instead propose an efficient homophily booster 
via a more careful design of ``shortcut'' edges. 

\subsection{Efficient Graph Homophily Booster}\label{ssec:method}
To address the computational inefficiency of the motivating na\"ive approach above, 
we propose an efficient, simple yet effective graph transformation method called \OursFull{} (\Ours) in this subsection. 

Note that the large number of NHB ``shortcut'' edges is because NHB \emph{directly} connects nodes with similar features. Since there are $O(|\CAL V|^2)$ node pairs in a graph, then the total number of added NHB ``shortcut'' edges can be as large as $O(|\CAL V|^2)$. 

To reduce the number of ``shortcut'' edges, we propose to connect such node pairs \emph{indirectly} instead. In particular, if we can create a few auxiliary ``hub'' nodes so that all such node pairs are \emph{indirectly} connected through the ``hub'' nodes, then we will be able to significantly reduce the number of ``shortcut'' edges at only a small price of adding a few ``hub'' nodes. Therefore, we need to develop an appropriate design of the ``hub'' nodes. 

\vspace{0.5em}\noindent\textbf{Graph transformation.} Following the aforementioned motivation, we propose to create a \emph{feature node} $x_k$ for each feature $k$ to serve as the ``hub'' nodes. Let $\CAL V_\CAL X$ denote the set of feature nodes:
\AL{\CAL V_\CAL X:=\{x_k:k\in\CAL X\}.}
To distinguish feature nodes $\CAL V_\CAL X$ from nodes $\CAL V$ in the original graph, we call $\CAL V$ \emph{graph nodes} from now on. For each graph node $v_i \in \CAL V$, if graph node $v_i$ has feature $k$ (i.e., $\BM X[v_i,k]=1$), we add an edge $(v_i,x_k)$ to connect the graph node $v_i$ and the feature node $x_k\in\CAL V_\CAL X$, and we call it a \emph{feature edge}. Let $\CAL E_\CAL X$ denote the set of feature edges:
\AL{\CAL E_\CAL X:={}&\{(v_i,x_k):v_i \in\CAL V,\,x_k \in\CAL V_\CAL X,\,\BM X[v_i,k]=1\}\nonumber\\\subseteq{}&\CAL V\times\CAL V_\CAL X.}
To distinguish feature edges $\CAL E_\CAL X$ from the edges $\CAL E$ in the original graph, we call $\CAL E$ \emph{graph edges} from now on.

Finally, we define the transformed graph $\CAL G^*=(\CAL V^*,\CAL E^*,\BM X^*)$ as follows. The nodes $\CAL V^*$ of the transformed graph $\CAL G^*$ are the original graph nodes $\CAL V$ and the added feature nodes $\CAL V_\CAL X$:
\AL{\CAL V^*:=\CAL V\cup\CAL V_\CAL X.}
The edges $\CAL E^*$ of the transformed graph $\CAL G^*$ are the original graph edges $\CAL E$ and the added feature edges $\CAL E_\CAL X$:
\AL{\CAL E^*:=\CAL E\cup\CAL E_\CAL X.}
We can also equivalently define the edges of the transformed graph $\CAL G^*$ by its adjacency matrix. Let $\BM A$ denote the adjacency matrix of the original graph $\CAL G$. Then, the adjacency matrix $\BM A^*$ of the transformed graph $\CAL G^*$ can be expressed in a block matrix form:
\AL{\BM A^*=\begin{bmatrix}
\BM A&\BM X\\
\BM X\Tp&\BM0
\end{bmatrix}.}

It remains to define node features $\BM X^*\in\BB R^{\CAL V^*\times\CAL X}$ of the transformed graph. For each graph node $v_i\in\CAL V$, we use 
its original features as its node features:
\AL{\BM X^*[v_i,:]:=\BM X[v_i,:].}
For each feature node $x_k\in\CAL V_\CAL X$, we define its node feature as the average feature vector among the graph nodes $v_i$ that are connected to feature node $x_k$:
\AL{\BM X^*[x_k,:]:=\frac{1}{|\CAL E_\CAL X\cap(\CAL V\times\{x_k\})|}\sum_{\begin{subarray}{c}v_i:(v_i,x_k)\in\CAL E_\CAL X\end{subarray}}\BM X[v_i,:].}

Our proposed graph transformation \Ours{} is illustrated in Figure~\ref{fig:illus}. In this example, $\{v_1,v_2,v_3,v_4,v_5\}$ are the graph nodes, where $v_1,v_2$ belong to one class, and $v_3,v_4,v_5$ belong to the other class. Our proposed \Ours{} adds feature nodes $x_1,x_2,x_3$ to the graph. We can see that feature node $x_1$ facilitates homophilic message passing between $v_1,v_2$, and that feature node $x_2$ facilitates homophilic message passing among $v_3,v_4,v_5$. 

\vspace{0.5em}\noindent\textbf{Theoretical guarantees.} The transformed graph $\CAL G^*$ enjoys a few theoretical guarantees. First, an important property of the feature edges is that every pair of nodes that share features can be connected through feature edges within two hops, as formally stated in Observation~\ref{lem:2hop}. This ensures that nodes with similar features are close to each other on the transformed graph $\CAL G^*$, facilitating homophilic message passing. 
\begin{OBS}[Two-hop indirect connection]\label{lem:2hop}
For each pair of nodes $u,v\in\CAL V$, if they share at least a feature (i.e., $\|\BM X[v_i,:]\land\BM X[v_j,:]\|_\infty>0$), then $v_i$ and $v_j$ are two-hop neighbors on the transformed graph $\CAL G^*$. 
\end{OBS}

Furthermore, we theoretically show that our proposed graph transformation \Ours{} can increase the homophily of the graph without increasing the size of the graph much, as formally stated in Theorem~\ref{thm:eff}. 

\begin{THM}[Efficient homophily booster]\label{thm:eff}
Given a heterophilic graph $\CAL G=(\CAL V,\CAL E,\BM X)$, let $\CAL G^*:=(\CAL V^*,\CAL E^*,\BM X^*)$ denote the graph transformed by our proposed \Ours{}. Under mild and realistic assumptions in Appendix~\ref{app:ass}, we have
\GA{
\OP{hom}(\CAL G^*)>\OP{hom}(\CAL G),\\
|\CAL V^*|\le O(|\CAL V|),\quad
|\CAL E^*|\le O(|\CAL E|).\label{eq:feat-edges}
}
\end{THM}
The effectiveness of our proposed \Ours{} is also empirically validated in Section~\ref{ssec:exp:homo}. As shown in Figure~\ref{fig:homo-analysis}, our proposed GRAPHITE significantly increases the homophily of originally heterophilic graph. 



\subsection{Neural Architecture}\label{ssec:method-detail}
The transformed graph $\CAL G^*$ can be readily fed into existing GNNs to boost their performance, even when the GNNs were originally designed for homophilic graphs, as demonstrated in Table~\ref{tab:abla}.  Meanwhile, to maximize the GNN performance on the transformed graph $\CAL G^*$, we introduce a GNN architecture specially designed for the transformed graph in this subsection.



To help the GNN distinguish graph nodes $\CAL V$ from feature nodes $\CAL V_\CAL X$, we use different edge weights for different edges. As a reference weight, suppose that graph edges $\CAL E$ have weight $w_\CAL E:=1$. Let 
$w_\CAL X>0$ denote the weight of feature edges $\CAL E_\CAL X$. Following GCN \cite{kipf2016semi}, we also use self-loops in GNN message passing; let $w_0>0$ denote the weight of self-loops. 

Let $d_u$ denote the weighted degree of each node $u\in\CAL V^*$. Specifically, for each graph node $v_i \in\CAL V$,
\AL{d_{v_i}:=w_0+\sum_{(v_i,v_j)\in\CAL E}w_{\CAL E}+\sum_{(v_i,x_k)\in\CAL E_\CAL X}w_\CAL X;}
and for each feature node $x_k\in\CAL V_{\CAL X}$,
\AL{d_{x_k}:=w_0+\sum_{(v_i,x_k)\in\CAL E_\CAL X}w_\CAL X.}

Inspired by FAGCN \cite{bo2021beyond}, we use a self-gating mechanism in GNN aggregation. For each node $u\in\CAL V^*$, let $\BM h_u\in\BB R^m$ denote the embedding of node $u$ before GNN aggregation, where $m$ is the embedding dimensionality. Then, the self-gating score $\alpha_{u,u'}$ between two nodes $u,u'\in\CAL V^*$ is defined as
\AL{\alpha_{u,u'}:=\tanh\Big(\frac{\BM a\Tp(\BM h_u\mathbin{\|}\BM h_{u'})+b}{\tau}\Big).}
where $\|$ denotes the concatenation operation, $\BM a\in\BB R^{2m}$ and $b\in\BB R$ are learnable parameters, and $\tau>0$ is a temperature hyperparameter. 

Next, we describe our aggregation mechanism. For each node $u\in\CAL V^*$, let $\BM h'_u\in\BB R^m$ denote the embedding of node $u$ after GNN aggregation. For each graph node $v_i \in\CAL V$, we define
\AL{\BM h'_{v_i}:={}&\frac{w_0\alpha_{v_i,v_i}}{\sqrt{d_{v_i}}\sqrt{d_{v_i}}}\BM h_{v_i}+\!\!\!\!\!\!\sum_{(v_i,v_j)\in\CAL E}\!\!\frac{\alpha_{v_i,v_j}}{\sqrt{d_{v_i}}\sqrt{d_{v_j}}}\BM h_{v_j}\nonumber\\&+\!\!\!\!\!\!\sum_{(v_i,x_j)\in\CAL E_\CAL X}\!\!\!\frac{w_\CAL X\alpha_{v_i,x_k}}{\sqrt{d_{v_i}}\sqrt{d_{x_k}}}\BM h_{x_k};}
and for each feature node $x_k\in\CAL V_\CAL X$, we define
\AL{\BM h'_{x_k}:=\frac{w_0\alpha_{x_k,x_k}}{\sqrt{d_{x_k}}\sqrt{d_{x_k}}}\BM h_{x_k}+\sum_{(v_i,x_k)\in\CAL E_\CAL X}\frac{w_\CAL X\alpha_{v_i,x_k}}{\sqrt{d_{v_i}}\sqrt{d_{x_k}}}\BM h_{v_i}.}
Furthermore, we add a multi-layer perceptron (MLP) with residual connections after each GNN aggregation. We use the GELU activation function \cite{hendrycks2016gaussian}.

\begin{table*}[t]
\centering
\caption{Summary of dataset statistics. We use four heterophilic graphs and two homophilic graphs.}
\label{tab:data}
\resizebox{0.65\linewidth}!{\begin{tabular}{l|cccc|cc}
\specialrule{3\arrayrulewidth}{5\arrayrulewidth}{5\arrayrulewidth}
\multirow{2}*{\textbf{Statistic}}&\multicolumn{4}{c|}{\textbf{Heterophilic Graphs}}&\multicolumn{2}{c}{\textbf{Homophilic Graphs}}\\
&\textsc{Actor}&\textsc{Squirrel-F}&\textsc{Chameleon-F}&\textsc{Minesweeper}&\textsc{Cora}&\textsc{CiteSeer}\\
\midrule
\# Nodes&7600&2223&890&10000&2708&3327\\
\# Edges&33544&46998&8854&39402&5429&4732\\
\# Features&931&2089&2325&7&1433&3703\\
\# Classes&5&5&5&2&7&6\\
Homophily & 0.0028 & 0.0086 & 0.0295 & 0.0094 & 0.7711 & 0.6707\\
\specialrule{2.4\arrayrulewidth}{5\arrayrulewidth}{5\arrayrulewidth}
\end{tabular}}
\vspace{-1em}
\end{table*}

\begin{table*}[t]
\centering
\caption{Comparison with existing methods. Our proposed \Ours{} \emph{significantly} outperforms state-of-the-art methods on heterophilic graphs while achieving \emph{comparable} accuracy with state-of-the-art methods on homophilic graphs.  Best results are marked in \textbf{bold}, and second best results are \underline{underlined}.
}
\label{tab:main-res}
\resizebox{0.95\linewidth}!{\begin{tabular}{l|cccc|cc}
\specialrule{2.4\arrayrulewidth}{5\arrayrulewidth}{5\arrayrulewidth}
\multirow{2}*{\textbf{Method}}&\multicolumn{4}{c|}{\textbf{Heterophilic Graphs}}&\multicolumn{2}{c}{\textbf{Homophilic Graphs}}\\
&\textsc{Actor}&\textsc{Squirrel-F}&\textsc{Chameleon-F}&\textsc{Minesweeper}&\textsc{Cora}&\textsc{CiteSeer}\\
\midrule
{MLP} & 35.04\PM1.53 & 33.91\PM1.55 & 38.44\PM5.14 & 50.99\PM1.47 & 75.45\PM1.88 & 71.53\PM0.70\\
\midrule
{ChebNet} & 34.40\PM1.18 & 31.75\PM3.42 & 34.30\PM4.33 & \underline{91.60}\PM0.44 & 81.58\PM5.09 & 65.18\PM8.37\\
{GCN} & 30.21\PM0.86 & 35.57\PM1.86 & 40.06\PM4.38 & 72.32\PM0.93 & 87.50\PM1.68 & 75.77\PM0.96\\
{SGC} & 29.26\PM1.41 & 38.27\PM2.16 & 41.40\PM4.91 & 72.11\PM0.95 & 88.05\PM2.08 & 75.80\PM1.75\\
{GAT} & 28.86\PM0.99 & 32.74\PM3.02 & 40.11\PM2.80 & 87.59\PM1.35 & 87.11\PM1.48 & 76.43\PM1.31\\
{GraphSAGE} & 34.95\PM1.06 & 34.43\PM2.68 & 39.33\PM4.53 & 90.54\PM0.66 & 87.90\PM1.73 & 76.43\PM1.19\\
{GIN} & 28.29\PM1.45 & 39.51\PM2.83 & 40.17\PM4.76 & 75.89\PM2.09 & 85.65\PM2.26 & 72.55\PM1.78\\
{APPNP} & 33.68\PM1.26 & 33.75\PM2.31 & 37.93\PM4.33 & 67.36\PM1.08 & 87.59\PM1.68 & 75.90\PM0.91\\
{GCNII} & 34.78\PM1.50 & 35.93\PM2.87 & 41.56\PM2.74 & 88.42\PM0.85 & 87.20\PM1.56 & 73.84\PM0.91\\
{GATv2} & 28.87\PM1.39 & 32.49\PM2.51 & 39.72\PM6.60 & 88.85\PM1.16 & 87.66\PM1.52 & 76.59\PM1.19\\
{MixHop} & 35.40\PM1.34 & 30.43\PM2.33 & 37.93\PM3.87 & 89.68\PM0.57 & 84.53\PM1.53 & 76.11\PM0.83\\
{TAGCN} & 34.92\PM1.19 & 33.33\PM2.37 & 41.01\PM3.77 & 91.54\PM0.56 & 88.38\PM1.95 & 76.49\PM1.41\\
{DAGNN} & 33.15\PM1.14 & 34.72\PM2.55 & 38.94\PM3.53 & 67.87\PM1.26 & 88.27\PM1.53 & 75.81\PM0.90\\
{JKNet} & 28.63\PM0.94 & \underline{40.81}\PM2.60 & 40.39\PM4.85 & 81.00\PM0.92 & 86.24\PM0.85 & 73.11\PM1.82\\
{Virtual Node} & 30.71\PM0.82 & 38.00\PM2.28 & 41.45\PM5.46 & 72.36\PM0.98 & 87.24\PM2.00 & 69.80\PM6.89\\
\midrule
{H2GCN} & 34.20\PM1.47 & 34.02\PM3.15 & 40.89\PM3.13 & 87.08\PM0.82 & 76.89\PM2.25 & 75.87\PM1.02\\
{FAGCN} & \underline{36.18}\PM1.52 & 36.52\PM1.72 & 39.83\PM3.93 & 84.69\PM2.05 & 88.66\PM2.11 & \underline{76.82}\PM1.48\\
{OrderedGNN} & 35.64\PM0.98 & 32.70\PM2.42 & 38.38\PM3.65 & 91.01\PM0.50 & 84.81\PM1.67 & 74.10\PM1.62\\
{GloGNN} & 19.80\PM2.61 & 28.72\PM2.63 & 40.17\PM4.66 & 53.42\PM1.47 & 73.02\PM2.98 & 72.46\PM2.09\\
{GGCN} & 32.76\PM1.39 & 35.06\PM5.65 & 34.08\PM3.44 & 84.76\PM1.84 & 86.39\PM1.93 & 75.36\PM1.99\\
{GPRGNN} & 35.42\PM1.33 & 34.97\PM2.83 & 40.50\PM4.55 & 83.94\PM0.98 & \textbf{88.86}\PM1.42 & 76.49\PM1.00\\
{ALT} & 33.10\PM1.38 & 37.28\PM1.49 & 39.61\PM3.36 & 89.06\PM0.64 & \underline{88.82}\PM2.02 & \textbf{76.88}\PM1.20\\
\midrule
{NodeFormer} & 29.26\PM2.31 & 24.29\PM2.60 & 34.92\PM4.08 & 77.71\PM3.50 & 87.44\PM1.37 & 75.20\PM1.27\\
{SGFormer} & 25.89\PM0.80 & 34.54\PM2.96 & \underline{42.79}\PM4.06 & 52.06\PM0.50 & 86.24\PM1.58 & 70.74\PM1.25\\
{DIFFormer} & 26.31\PM1.19 & 33.17\PM2.84 & 39.16\PM4.10 & 69.25\PM0.93 & 86.61\PM3.04 & 76.65\PM1.52\\
\midrule
\textbf{\Ours{} }(Ours)&\textbf{37.69}\PM1.57&\textbf{43.06}\PM2.89&\textbf{45.08}\PM4.04&\textbf{94.78}\PM0.41&88.23\PM1.65&76.41\PM1.57\\
\specialrule{3\arrayrulewidth}{5\arrayrulewidth}{5\arrayrulewidth}
\end{tabular}}
\end{table*}

\section{Experiments}

We conduct extensive experiments on both heterophilic and homophilic datasets to answer the following research questions:
\begin{enumerate}
\renewcommand\labelenumi{\textbf{RQ\theenumi:}\!\!\!\!\!\!\!\!\!}
\item\label{rq:main}\,\,\,\,\,\,\,\,\,How does the proposed framework \Ours{} compare with state-of-the-art methods? 
\item\label{rq:homo}\,\,\,\,\,\,\,\,\,How much improvement can the proposed graph transformation achieve in the graph homophily?
\item\label{rq:abla}\,\,\,\,\,\,\,\,\,Can the proposed graph transformation alone enhance the performance of existing homophilic GNNs?
\end{enumerate}

\subsection{Experimental Settings}

\vspace{0.5em}\noindent\textbf{Datasets.}
We evaluate \Ours{} and various baseline methods across six real-world datasets, with their statistics summarized in Table~\ref{tab:data}. The reported homophily is the \textit{adjusted homophily} introduced in~\cite{platonov2024characterizing}, which exhibits more desirable properties compared to traditional edge/node homophily. We leverage \textit{adjusted homophily} to categorize the datasets into two groups: \textit{heterophilic} and \textit{homophilic}. Please see the appendix for dataset descriptions.


\begin{table}[t]
\caption{Effectiveness of the proposed graph transformation. \Ours{} transformed graphs alone can already enhance the performance of homophilic GNNs.
}
\label{tab:abla}
\centering
\resizebox{\linewidth}{!}{\begin{tabular}{l|cc|cc}
\toprule
\textbf{Dataset} & \multicolumn{2}{c|}{\textsc{Actor}}  & \multicolumn{2}{c}{\textsc{Minesweeper}}  \\
\small{+\Ours{}}?&\xmark&\cmark&\xmark&\cmark\\
\midrule
GCN      &30.21\PM0.86&\textbf{34.83}\PM1.28&72.32\PM0.93&\textbf{75.38}\PM1.56\\
GAT      &28.86\PM0.99&\textbf{32.09}\PM1.35&87.59\PM1.35&\textbf{88.66}\PM0.88\\
GraphSAGE&34.95\PM1.06&\textbf{35.09}\PM1.06&90.54\PM0.66&\textbf{90.85}\PM0.67\\
JKNet    &28.63\PM0.94&\textbf{35.96}\PM1.40&81.00\PM0.92&\textbf{85.56}\PM2.59\\
GIN      &28.29\PM1.45&\textbf{33.75}\PM1.83&75.89\PM2.09&\textbf{87.07}\PM1.71\\
\bottomrule
\end{tabular}}
\vspace{-1em}
\end{table}

\begin{table}[t]
\caption{Relative improvement ratio of \textit{feature homophily} and \textit{adjusted homophily} across datasets. Larger values represent more significant homophily boost after applying \Ours{}. See Figure~\ref{fig:homo-analysis} for visualization.}
\label{tab:homo-ratio-increment}
\centering
\begin{tabular}{lcc}
\toprule
\textbf{Dataset} & $\Delta H^{\textnormal{feature}}(\mathcal{G})$   & $\Delta H^{\textnormal{adj}}(\mathcal{G})$      \\
\midrule
\textsc{Actor}       &    2.79      &   28.67     \\
\textsc{Squirrel-F}  &    10.61      &   3.15 \\
\textsc{Chameleon-F} &    18.39      &  5.02  \\
\textsc{Minesweeper} &    1.41      &   11.23 \\
\bottomrule
\end{tabular}
\vspace{-1em}
\end{table}

\begin{table*}[t]
\caption{Summary of baseline methods.}
\label{tab:baselines}
\centering
\resizebox{0.8\linewidth}{!}{
\begin{tabular}{c|c}
\toprule
\textbf{Type} & \textbf{Baseline Methods}  \\
\midrule
\makecell{Non-Graph} & \makecell[l]{Multi-Layer Perceptron (MLP)}  \\
\midrule
\makecell{Homophilic\\GNNs}    & \makecell[l]{ChebNet~\cite{defferrard2016convolutional}, GCN~\cite{kipf2016semi}, SGC~\cite{wu2019simplifying},\\GAT~\cite{velivckovic2018graph},  GraphSAGE~\cite{hamilton2017inductive}, GIN~\cite{xu2018powerful},\\APPNP~\cite{gasteiger2018predict}, GCNII~\cite{chen2020simple}, GATv2~\cite{brody2021attentive},\\MixHop~\cite{abu2019mixhop}, TAGCN~\cite{du2017topology}, DAGNN~\cite{liu2020towards},\\JKNet~\cite{xu2018representation}, Virtual Node~\cite{gilmer2017neural}} \\
\midrule

\makecell{Heterophilic\\GNNs}  & \makecell[l]{ H2GCN~\cite{zhu2020beyond}, FAGCN~\cite{bo2021beyond}, OrderedGNN~\cite{song2023ordered}, GloGNN~\cite{li2022finding}, \\GGCN~\cite{yan2022two}, GPRGNN~\cite{chien2020adaptive}, ALT~\cite{xu2023node}}  \\
\midrule
\makecell{Graph\\Transformers}  & \makecell[l]{NodeFormer~\cite{wu2022nodeformer}, SGFormer~\cite{wu2024simplifying}, DIFFormer~\cite{wu2023difformer}} \\
\bottomrule
\end{tabular}
}
\end{table*}

\vspace{0.5em}\noindent\textbf{Baseline methods.} 
In our experiments, we consider a wide range of GNN baselines, including MLP (structure-agnostic), homophilic GNNs, heterophilic GNNs, and Graph Transformers. The full list is shown in Table~\ref{tab:baselines}. Please see the appendix for descriptions of baseline methods.

\vspace{0.5em}\noindent\textbf{Training and evaluation.} To benchmark \Ours{} and compare it with the baseline methods, we use \emph{node classification} tasks with performance measured by classification accuracy on Actor, Chameleon-Filtered (Chameleon-F), Squirrel-Filtered (Squirrel-F), Cora, and CiteSeer and by ROC-AUC on Minesweeper following \cite{platonov2023critical}. For all baseline methods, we use the hyperparameters provided by the authors. For the evaluation of Actor, Chameleon-F and Squirrel-F, we generate 10 random splits with a ratio of $48\%/32\%/20\%$ as the training/validation/test set, respectively, following \cite{gu2024universal}. For the evaluation of Minesweeper, we directly utilize the 10 random splits provided by the original paper \cite{platonov2023critical}. For the evaluation of Cora and CiteSeer, we follow~\cite{luan2021heterophily, chien2020adaptive} to randomly generate 10 random splits with a ratio of $60\%/20\%/20\%$ as the training/validation/test set, respectively. For each experiment, we report the mean and the standard deviation of the performance metric across the corresponding 10 random splits. Please see the appendix for additional experimental settings.




\vspace{0.5em}\noindent

\subsection{Main Results}

To answer RQ\ref{rq:main}, we compare the proposed method \Ours{} with 25 state-of-the-art methods on six heterophilic and homophilic graphs. The results are shown in Table~\ref{tab:main-res}.





As shown in Table~\ref{tab:main-res}, \Ours{} achieves significant performance gains (p-value$<$0.1) over prior state-of-the-art GNN methods on heterophilic graphs while maintaining competitive accuracy on homophilic graphs. Specifically, \Ours{} outperforms the best baseline methods by $4.17\%, 5.23\%, 5.35\%$ and $3.47 \%$ on \textsc{Actor}, \textsc{Squirrel-F}, \textsc{Chemeleon-F} and \textsc{Minesweeper}, respectively. While some existing models perform well on individual datasets, they often struggle on others, highlighting their insufficient consistency. In contrast, \Ours{} demonstrates the best results across all four heterophilic benchmarks. Another interesting observation is that while \Ours{} is built upon FAGCN~\cite{bo2021beyond}, it significantly surpasses FAGCN, demonstrating the effectiveness of the beneficial effect of graph transformation and \textit{feature edges}.

\vspace{0.5em}\noindent\textbf{Discussion.} It is worth noting that most of the baseline methods cannot achieve better results compared to MLP on \textsc{Actor}, which can be explained by the fact that these methods typically treat node features and graph structure as joint input without explicitly decoupling them. The weak structural homophily exhibited by \textsc{Actor} makes typical GNNs fail to capture important feature signals, reinforcing the importance of our graph transformation strategy that boosts \textit{feature homophily} significantly. For \textsc{Squirrel-F}, we find that JKNet is the best among baselines. This observation reveals that structure information is very important within \textsc{Squirrel-F} since JKNet aggregates feature knowledge from multi-hop neighbors to learn structure-aware representation. This finding also explains the success of \Ours{} since the useful multi-hop information in \textsc{Squirrel-F} can be propagated even more efficiently through the constructed \textit{feature edges}. 

As another example, SGFormer performs the best on \textsc{Chameleon-F} among baseline methods. We argue that \textsc{Chameleon-F} needs a considerable amount of global messages and graph transformers are experts at capturing this type of information. Compared with NodeFormer and DIFFormer, SGFormer is the most advanced graph transformer utilizing simplified graph attention that strikes a good balance between global structural information and feature signal, preventing the over-globalizing issue \cite{xing2024less}. Similarly, \Ours{} transforms the original graph into a form that facilitates global message exchange by the introduction of \textit{feature edges}. As a final remark, although \Ours{} is designed specifically to deal with heterophilic datasets, \Ours{} still maintains competitive accuracy on homophilic datasets (\textsc{Cora} and \textsc{CiteSeer}), achieving results that are on par with the best existing methods. 

%


\subsection{Homophily Analysis}\label{ssec:exp:homo}
To answer RQ\ref{rq:homo}, we conduct a homophily analysis across heterophilic datasets under two homophily metrics: \textit{feature homophily} and \textit{adjusted homophily}, whose formal definition can be found in Appendix~\ref{app:homo-metric}. As shown in Figure~\ref{fig:homo-analysis}, we can observe a significant boost in both two homophilily metrics after applying \Ours{} across heterophilic datasets. The relative improvement ratio is presented in Table~\ref{tab:homo-ratio-increment}, where $\Delta H^{\cdot} (\mathcal{G})$ is the ratio between the corresponding homophily metric computed on original graph and the graph after applying \Ours{}. 

\vspace{0.5em}\noindent\textbf{Discussion.} Overall, \Ours{} 
effectively boosts both homophily metrics across all heterophilic datasets. Specifically, Squirrel-F and Chameleon-F demonstrate significant boosts in terms of \textit{feature homophily}. This is mainly because their discrete features directly correspond to specific topics and each feature edge will contribute much higher feature similarity than usual edges. On the other hand, Actor and Minesweeper showcase much higher \textit{adjusted homophily} after applying \Ours{}. For Actor, this favorable behavior can be attributed to the high correlation between page co-occurrences and node labels; while for Minesweeper, the sum of label-specific node degrees (defined in Equation~(\ref{eq:adj-homo})) increases much due to the transformation performed by \Ours{}.


\subsection{Ablation Studies}\label{ssec:exp:abla}
To further demonstrate the effectiveness of our proposed graph transformation \Ours{} and answer RQ\ref{rq:abla}, we compare the performance of homophilic GNNs on the original graph and that on the transformed graph. In this experiment, we use two larger-scale datasets, \textsc{Actor} and \textsc{Minesweeper}, and five representative homophilic GNNs, GCN, GAT, GraphSAGE, JKNet, and GIN. The results are presented in Table~\ref{tab:abla}. 

From Table~\ref{tab:abla}, we can see that our proposed \Ours{} consistently improves the performance of the five representative homophilic GNNs on both datasets, even though these GNNs are not specially designed for modeling feature nodes. For example, the accuracy of GAT on \textsc{Actor} is enhanced from 30.21\% to 34.83\%, which is a relative improvement of 15.29\%. The results demonstrate that our proposed graph transformation \Ours{} can significantly enhance the performance of homophilic GNNs on originally heterophilic graphs, echoing the fact that our proposed graph transformation can significantly increase the graph homophily. 



\section{Related Work}




\vspace{0.5em}\noindent\textbf{Heterophily.} A substantial body of research has explored the challenges of heterophily in graph neural networks (GNNs). Many early approaches sought to improve information aggregation, such as MixHop~\cite{abu2019mixhop}, which mixes different-hop neighborhood features, and GPRGNN~\cite{chien2020adaptive}, which employs generalized PageRank propagation for adaptive message passing. Other methods focus on explicit heterophilic adaptations, such as H2GCN~\cite{zhu2020beyond}, which separates ego- and neighbor-embeddings, and FAGCN~\cite{bo2021beyond}, which learns optimal representations via frequency-adaptive filtering. Additional works, including OrderedGNN~\cite{song2023ordered}, GloGNN~\cite{li2022finding}, and GGCN~\cite{yan2022two}, leverage structural ordering, global context, and edge corrections, respectively, to enhance performance on heterophilic graphs. Recent advances explore alternative formulations, such as component-wise signal decomposition (ALT~\cite{xu2023node}) and adaptive residual mechanisms~\cite{xuslog, yan2024trainable} for greater flexibility. Beyond architectural innovations, rigorous benchmarking efforts~\cite{lim2021large, zhu2024impact, platonov2023critical} have been introduced to standardize evaluations and assess generalization across diverse graph properties. A broader synthesis of heterophilic GNN techniques can be found in recent surveys~\cite{zheng2022graph, zhu2023heterophily, luan2024heterophilic, gong2024survey}.


\vspace{0.5em}\noindent\textbf{Over-squashing.} 
A problem related to heterophily is over-squashing.
The over-squashing problem in Message Passing Neural Networks (MPNNs) arises when long-range information is exponentially compressed, preventing effective dissemination across the graph~\cite{alon2020bottleneck, shi2023exposition}. A primary research direction addresses this issue by identifying topological bottlenecks and modifying graph connectivity. \cite{topping2021understanding} established an initial framework linking oversquashing to graph Ricci curvature, demonstrating that negatively curved edges act as bottlenecks. Building on this idea, subsequent works have developed rewiring strategies inspired by curvature-based principles~\cite{nguyen2023revisiting, shi2023curvature}. Beyond curvature, \cite{black2023understanding} introduced a perspective using effective resistance.
Another line of research leverages spectral methods to counteract over-squashing, with notable approaches including spectral gaps~\cite{arnaiz2022diffwire}, expander graph constructions~\cite{deac2022expander}, and first-order spectral rewiring~\cite{karhadkar2022fosr}. More recently, \cite{di2023over} provided a comprehensive analysis of the factors contributing to oversquashing. Additional solutions explore advanced rewiring strategies and novel message-passing paradigms~\cite{barbero2023locality,qian2023probabilistically,behrouz2024graph}.

\nocite{wei2024robust,bao2025latte,chen2024wapiti,liu2025breaking,liu2024logic,liu2024class,liu2024aim,liu2023topological,zeng2025pave,zeng2024graph,lin2025moralise,lin2024backtime,qiu2025saffron,qiu2025ask,qiu2025efficient,qiu2024tucket,qiu2023reconstructing,qiu2022dimes,xu2024discrete,li2025model,zou2025transformer,qiu2024gradient,yoo2025embracing,yoo2025generalizable,yoo2024ensuring,chan2024group,wu2024fair,he2024sensitivity,wang2023networked}

\section{Conclusion}

In this paper, we propose \Ours{}, a simple yet efficient framework to address the heterophily issue in node classification. By introducing feature nodes that connect to graph nodes with corresponding discrete features, we can solve the heterophily issue by increasing the graph homophily ratio. 
Through theoretical analysis and empirical study, we validate that \Ours{} can indeed effectively increase the graph homophily. Our extensive experiments demonstrate that \Ours{} consistently outperforms state-of-the-art methods, achieving significant performance gains on heterophilic graph datasets and comparable performance on homophilic graph datasets. An interesting future direction would be extending the proposed graph transformation to general graphs with continuous node features. 

\bibliography{AAAI26/80-Reference}

\appendix
\section{Experimental Settings (Cont'd)}
\subsection{Datasets (Cont'd)}
For heterophilic group, we consider the following datasets, which are widely used as benchmarks for studying graph learning methods under heterophilic settings.
\begin{itemize}
\setlength\itemsep{0.5em}
    \item \textsc{Actor}~\cite{pei2020geom}: Actor dataset is an actor-only induced subgraph of the film dataset introduced by~\cite{tang2009social}. The nodes are actors and the edges denote co-occurrence on the same Wikipedia page. The node features are keywords on the pages and we classify nodes into five categories. 
    \item \textsc{Squirrel-F}~\cite{platonov2023critical}: Squirrel-Filtered (Squirrel-F) is a page-page dataset. It is a subset of the Wiki dataset~\cite{rozemberczki2021multi} that focus on the topic related to squirrel. Nodes are web pages and edges are mutual links between pages. The node features are important keywords in the pages and we classify nodes into five categories in terms of traffic of the webpage. 
    
    \item \textsc{Chameleon-F}~\cite{platonov2023critical}: Chameleon-Filtered (Chameleon-F) is a page-page dataset. It is a subset of the Wiki dataset~\cite{rozemberczki2021multi} that focus on the topic related to chameleon.  Nodes are web pages and edges are mutual links between pages. The node features are important keywords in the pages and we classify nodes into five categories in terms of traffic of the webpage. 
    
    \item \textsc{Minesweeper}~\cite{platonov2023critical}: Minesweeper dataset is a synthetic dataset that simulates a Minesweeper game with 100x100 grid. Each node is connected to its neighboring nodes where $20\%$ nodes are selected as mines at random. Node features are numbers of neighboring mines and the goal is to predict whether each test node is mine. These datasets are widely used as benchmarks for studying graph learning methods under heterophilic settings.
\end{itemize}
    For the homophilic group, we consider the following datasets, which are standard homophilic network benchmarks.
\begin{itemize}

    \item \textsc{Cora}~\cite{sen2008collective} : Cora dataset is a citation network, where nodes represent scientific papers in the machine learning field, and edges correspond to citation relationships between these papers. Each node is associated with a set of features that describe the paper, represented as a bag-of-words model. The task for this dataset is to classify each paper into one of seven categories, reflecting the area of research the paper belongs to.
    
    \item \textsc{CiteSeer}~\cite{sen2008collective}:
    CiteSeer dataset is a citation network of scientific papers. It consists of research papers as nodes, with citation links forming the edges between them. Each node is associated with a set of features derived from the paper's content, which is a bag-of-words representation of the paper's text. The task for this dataset is to classify each paper into one of six categories, each representing a specific field of study. 
\end{itemize}

\subsection{Baseline Methods (Cont'd)}
We briefly introduce GNN-based baseline methods as follows. 

The first category is \textit{homophilic GNNs}, which are originally designed under the homophily assumption.
\begin{itemize}
    \item ChebNet~\cite{defferrard2016convolutional}: Uses Chebyshev polynomials to approximate graph convolutions.
    \item GCN~\cite{kipf2016semi}: Employs a first-order Chebyshev approximation for spectral graph convolutions.
    \item SGC~\cite{wu2019simplifying}: Simplifies GCN by removing non-linearities and collapsing weight matrices for efficiency.
    \item GAT~\cite{velivckovic2018graph}: Introduces attention mechanisms to assign adaptive importance to edges.
    \item GraphSAGE~\cite{hamilton2017inductive}: Uses several aggregators for inductive graph learning.
    \item GIN~\cite{xu2018powerful}: Employs sum-based aggregation to maximize graph structure expressiveness.
    \item APPNP~\cite{gasteiger2018predict}: Combines personalized PageRank with neural propagation.
    \item GCNII~\cite{chen2020simple}: Extends GCN with residual connections and identity mapping for deep GNN training.
    \item GATv2~\cite{brody2021attentive}: Enhances GAT with dynamic attention coefficients for flexible neighbor weighting.
    \item MixHop~\cite{abu2019mixhop}: Aggregates multi-hop neighborhood features by mixing different powers of adjacency matrices.
    \item TAGCN~\cite{du2017topology}: Introduces trainable polynomial filters for adaptive, multi-scale feature extraction.
    \item DAGNN~\cite{liu2020towards}: Uses dual attention to decouple message aggregation and transformation, improving depth scalability.
    \item JKNet~\cite{xu2018representation}: Uses a jumping knowledge mechanism to combine features from different layers adaptively. We default the backbone GNN model to GCN.
    \item Virtual Node~\cite{gilmer2017neural}: Introduces an auxiliary global node to facilitate message passing. We default the backbone GNN model to GCN.
\end{itemize}

The second category is \textit{heterophilic GNN}s, which are designed for graphs where connected nodes often have different labels.

\begin{itemize}
    \item H2GCN~\cite{zhu2020beyond}: Enhances GNNs by ego-/neighbor-embedding seperation, higher-order neighbors and intermediate representation combinations.
    \item FAGCN~\cite{bo2021beyond}: Uses frequency adaptive filtering to learn optimal graph representations.
    \item OrderedGNN~\cite{song2023ordered}: Aligns the order to encode neighborhood information and avoids feature mixing.
    \item GloGNN~\cite{li2022finding}: Incorporates global structural information to enhance graph learning beyond local neighborhoods.
    \item GGCN~\cite{yan2022two}: Utilizes structure/feature-based edge correction to combat over-smoothing and heterophily.
    \item GPRGNN~\cite{chien2020adaptive}: Introduces generalized PageRank propagation to  capture the graph structure.
    \item ALT~\cite{xu2023node}: Decomposes graph into components, extracts signals from these components and adaptively integrate these signals.
\end{itemize}

The last category is \textit{graph transformers}, which adapt transformer architectures to graph data and look beyond local neighborhood aggregation. 
\begin{itemize}
    \item NodeFormer~\cite{wu2022nodeformer}:  Introduces all-pair message passing on layer-specific adaptive latent graphs, enabling global feature propagation with linear complexity.
    \item SGFormer~\cite{wu2024simplifying}:  Develops a graph encoder backbone that efficiently computes all-pair interactions with one-layer attentive propagation.
    \item DIFFormer~\cite{wu2023difformer}:
    Proposes an energy-constrained diffusion model, leading to variants that are efficient and capable of capturing complex structures.
\end{itemize}



\subsection{Training \& Evaluation (Cont'd)}
For our method, we use $w_\CAL X\in\{0.01,0.1,0.6,8\}$, $w_0\in\{0.1,0.2,0.3,0.5,1,8\}$, $\tau\in\{0.01,0.1,1\}$, and dropout rate 0.2. We use the GNN architecture described in the method section 
with 8 GNN layers with hidden dimensionality 512 and add a two-layer MLP after each GNN layer for heterophilic graphs and use FAGCN for homophilic graphs. We use original node features as described in Section~\ref{ssec:method}, except that we use zeros as the features of graph nodes on Squirrel-F and that we normalize the features of graph nodes on Cora and CiteSeer after computing the features of feature nodes. We train the GNN with learning rate 0.00003 for 1000 steps using the Adam optimizer \cite{kingma2014adam}. Experiments were implemented in PyTorch 2.7.0 and Deep Graph Library (DGL) 2.4.0 and were run on Intel Xeon CPU @ 2.20GHz with 96GB memory and NVIDIA Tesla V100 32GB GPU.

\section{Definition of Homophily Metrics}
\label{app:homo-metric}

To measure to what extent \Ours{} can  boost graph homophily on heterophlic datasets, we consider two popular homophily metrics: \textit{feature homophily}~\cite{jin2022raw} and \textit{adjusted homophily}~\cite{platonov2024characterizing}. Formally, given a graph $\mathcal{G}, $\textit{feature homophily} $H^{\textnormal{feature}}$ is defined as follows:
\begin{equation}
\label{eq:homo-equation}
    H^{\textnormal{feature}} (\mathcal{G}) := \frac1{|\mathcal{E}|}\sum_{(v_i,v_j)\in \mathcal{E}} \textnormal{sim}(v_i,v_j),
\end{equation}
where $\textnormal{sim}(v_i,v_j) := \OP{cos}(\BM X[v_i,:], \BM X[v_j,:])$ is the cosine-similarity computed between features of nodes $v_i,v_j$. This metric is a variant of the \textit{generalized edge homophily ratio} $H^{\textnormal{edge}}$ proposed by~\cite{jin2022raw}, which measures the feature similarity between each of the connected node pairs in the graph dataset. Then, the \textit{adjusted homophily} ($H^{\textnormal{adj}}(\mathcal{G})$) is defined as follows:
\begin{equation}
    H^{\textnormal{adj}}(\mathcal{G}) := \frac{H^{\textnormal{edge}}(\mathcal{G}) - \sum_{k=1}^C D_c^2/(2|\mathcal{E}|)^2}{1 - \sum_{c=1}^C D_c^2/(2|\mathcal{E}|)^2},
\label{eq:adj-homo}
\end{equation}
where $C$ denotes the number of classes and $H^{\textnormal{edge}}(\mathcal{G})$ is \textit{edge homophily}, which is defined similarly as Equation~(\ref{eq:homo-equation}) with the similarity function $\textnormal{sim}(v_i,v_j) = \mathbf{1}_{\{y_{v_i}=y_{v_j}\}}$, and
\AL{D_c := \sum_{v:y_v=c} \OP{deg}(v)\label{eq:node-degree-sum}}
is the sum of node degrees with a specific node label $c$. Note that $y_{v_i}$ stands for the node label of graph node $v_i$. Since we do not have node labels for the \textit{feature nodes} when computing \textit{adjusted homophily}, we assign them ``soft label'', which is a uniform probability distribution over classes, obtained by aggregating the labels of its 1-hop neighbors.

\section{Theoretical Analysis}\label{app:proofs}

\subsection{Assumptions}\label{app:ass}
In this subsection, we introduce the assumptions of our theoretical analysis, which are mild and realistic. 

Given a graph $\CAL G=(\CAL V,\CAL E,\BM X)$ with $\CAL E\ne\varnothing$ and $\BM X\in\{0,1\}^{\CAL V\times\CAL X}$, we define the feature similarity metric as $\OP{sim}(v_i,v_j):=\|\BM X[v_i,:]\land\BM X[v_j,:]\|_\infty$ and use the feature homophily as the homophily metric:
\AL{\OP{hom}(\CAL G):=\frac1{|\CAL E|}\sum_{(v_i,v_j)\in\CAL E}\OP{sim}(v_i,v_j).}
Furthermore, we assume that the original graph $\CAL G$ is heterophilic. That is, we have $\OP{hom}(\CAL G)<1$ while there exists a pair of nodes, $v_i,v_j\in\CAL V$ ($v_i\ne v_j$), such that $\OP{sim}(v_i,v_j)>0$ but $(v_i,v_j)\notin\CAL E$. 

Besides that, we assume that the given graph $\CAL G$ does not have too dense features. Formally, we assume that $|\CAL X|\le O(|\CAL V|)$ and that $\|\BM X\|_0\le O(|\CAL E|)$.
For the transformed graph $\CAL G^*$, we assume that every feature is used: for any feature $k\in\CAL X$, there exists a graph node $v_i\in\CAL V$ such that $\BM X[v_i,k]=1$. 

\subsection{Technical Lemma}
Here, we prove a technical lemma that we will use later.

\begin{LEM}\label{lem:add}
Let $\CAL A,\CAL B\subset\BB R$ be two nonempty, finite multisets with $z'>\frac1{|\CAL A|}\sum_{z\in\CAL A}z$ for all $z'\in\CAL B$. Then,
\AM{\frac{1}{|\CAL A\sqcup\CAL B|}\sum_{z\in\CAL A\sqcup\CAL B}z>\frac1{|\CAL A|}\sum_{z\in\CAL A}z.}
\end{LEM}

\begin{proof}
To simplify notation, let
\AL{
\mu&:=\frac1{|\CAL A|}\sum_{z\in\CAL A}z,\\
\varDelta&:=\min\CAL B-\mu>0.
}
Then,
\AL{
&\frac{1}{|\CAL A\sqcup\CAL B|}\sum_{z\in\CAL A\sqcup\CAL B}z-\frac1{|\CAL A|}\sum_{z\in\CAL A}z
\\={}&\frac{1}{|\CAL A|+|\CAL B|}\bigg(\sum_{z\in\CAL A}z+\sum_{z\in\CAL B}z\bigg)-\frac1{|\CAL A|}\sum_{z\in\CAL A}z
\\={}&\frac{1}{|\CAL A|+|\CAL B|}\bigg(|\CAL A|\cdot\frac1{|\CAL A|}\sum_{z\in\CAL A}z+\sum_{z\in\CAL B}z\bigg)-\frac1{|\CAL A|}\sum_{z\in\CAL A}z
\\={}&\frac{1}{|\CAL A|+|\CAL B|}\bigg(|\CAL A|\cdot\mu+\sum_{z\in\CAL B}z\bigg)-\mu
\\\ge{}&\frac{1}{|\CAL A|+|\CAL B|}\bigg(|\CAL A|\cdot\mu+\sum_{z\in\CAL B}\min\CAL B\bigg)-\mu
\\={}&\frac{1}{|\CAL A|+|\CAL B|}\big(|\CAL A|\cdot\mu+|\CAL B|\cdot\min\CAL B\big)-\mu
\\={}&\frac1{|\CAL A|+|\CAL B|}\big(|\CAL B|\cdot\min\CAL B-|\CAL B|\cdot\mu\big)
\\={}&\frac{|\CAL B|}{|\CAL A|+|\CAL B|}(\min\CAL B-\mu)
\\={}&\frac{|\CAL B|}{|\CAL A|+|\CAL B|}\varDelta>0
.}
It follows that
\AL{\frac{1}{|\CAL A\sqcup\CAL B|}\sum_{z\in\CAL A\sqcup\CAL B}z&>\frac1{|\CAL A|}\sum_{z\in\CAL A}z.\qedhere}
\end{proof}

\subsection{Proof of Theorem~\ref{thm:naive}}
\vspace{0.5em}\noindent\textbf{Homophily.} Since the original graph $\CAL G$ is homophilic, then there exists a pair of nodes, $v_i,v_j\in\CAL V$ ($v_i\ne v_j$), such that $\OP{sim}(v_i,v_j)=\|\BM X[v_i,:]\land\BM X[v_j,:]\|_\infty>0$ but $(v_i,v_j)\notin\CAL E$. According to the definition of $\CAL E^\dagger$, we know that $(v_i,v_j)\in\CAL E^\dagger\setminus\CAL E\ne\varnothing$, so $\CAL E^\dagger\setminus\CAL E\ne\varnothing$. 

Furthermore, for any $(v_i,v_j)\in\CAL E^\dagger\setminus\CAL E$, since $\OP{sim}(v_i,v_j)=\|\BM X[v_i,:]\land\BM X[v_j,:]\|_\infty>0$, then there exists a feature $k\in\CAL X$ such that $\BM X[v_i,k]\land\BM X[v_j,k]>0$. Since the feature matrix $\BM X$ is binary, then we must have
\AL{\BM X[v_i,k]=1,\qquad\BM X[v_j,k]=1.}
It follows that
\AL{\OP{sim}(v_i,v_j)&=\|\BM X[v_i,:]\land\BM X[v_j,:]\|_\infty\\
&=\max_{k'\in\CAL X}|\BM X[v_i,k']\land\BM X[v_j,k']|\\
&\ge|\BM X[v_i,k]\land\BM X[v_j,k]|\\
&=|1\land1|=1
.}
Since $\OP{hom}(\CAL G)<1$, then
\AL{\OP{sim}(v_i,v_j)\ge1>\OP{hom}(\CAL G).}
Therefore, by Lemma~\ref{lem:add} with
\AL{
\CAL A&:=\{\!\!\{\OP{sim}(v_i,v_j):(v_i,v_j)\in\CAL E\}\!\!\},\\
\CAL B&:=\{\!\!\{\OP{sim}(v_i,v_j):(v_i,v_j)\in\CAL E^\dagger\setminus\CAL E\}\!\!\},
}
we have
\AL{
\OP{hom}(\CAL G^\dagger)&=\frac1{|\CAL E^\dagger|}\sum_{(v_i,v_j)\in\CAL E^\dagger}\OP{sim}(v_i,v_j)\\
&=\frac1{|\CAL E\sqcup(\CAL E^\dagger\setminus\CAL E)|}\sum_{(v_i,v_j)\in\CAL E\sqcup(\CAL E^\dagger\setminus\CAL E)}\OP{sim}(v_i,v_j)\\
&=\frac{1}{|\CAL A\sqcup\CAL B|}\sum_{z\in\CAL A\sqcup\CAL B}z\\
&>\frac1{|\CAL A|}\sum_{z\in\CAL A}z\\
&=\frac1{|\CAL E|}\sum_{(v_i,v_j)\in\CAL E}\OP{sim}(v_i,v_j)\\
&=\OP{hom}(\CAL G)
.}

\vspace{0.5em}\noindent\textbf{Number of edges.} Since there are $|\CAL V|$ nodes in total, then the total number of node pairs is $\binom{|\CAL V|}2$. Recall that $\CAL E^\dagger\setminus\CAL E$ is the set of added edges. It follows that
\AL{|\CAL E^\dagger|-|\CAL E|&=|\CAL E^\dagger\setminus\CAL E|\le\binom{|\CAL V|}2\\&=\frac{|\CAL V|(|\CAL V|-1)}2=O(|\CAL V|^2).}

\subsection{Proof of Observation~\ref{lem:2hop}}
Since $\OP{sim}(v_i,v_j)=\|\BM X[v_i,:]\land\BM X[v_j,:]\|_\infty>0$, then there exists a feature $k\in\CAL X$ such that $\BM X[v_i,k]\land\BM X[v_j,k]>0$. Since the feature matrix $\BM X$ is binary, then we must have
\AL{\BM X[v_i,k]=1,\qquad\BM X[v_j,k]=1.}
This implies that $(v_i,x_k)\in\CAL E^*$ and that $(v_j,x_k)\in\CAL E^*$. Hence, there exists a length-$2$ path $v_i\to x_k\to v_j$ connecting graph nodes $v_i$ and $v_j$. Therefore, $v_i$ and $v_j$ are two-hop neighbors of each other. 

\subsection{Proof of Theorem~\ref{thm:eff}}
\vspace{0.5em}\noindent\textbf{Homophily.} Since the original graph $\CAL G$ is homophilic, then there exists a pair of nodes, $v_i,v_j\in\CAL V$ ($v_i\ne v_j$), such that $\OP{sim}(v_i,v_j)=\|\BM X[v_i,:]\land\BM X[v_j,:]\|_\infty>0$ but $(v_i,v_j)\notin\CAL E$. Since $\OP{sim}(v_i,v_j)=\|\BM X[v_i,:]\land\BM X[v_j,:]\|_\infty>0$, then there exists a feature $k\in\CAL X$ such that $\BM X[v_i,k]\land\BM X[v_j,k]>0$. Since the feature matrix $\BM X$ is binary, then we must have
\AL{\BM X[v_i,k]=1,\qquad\BM X[v_j,k]=1.}
This implies that $(v_i,x_k)\in\CAL E^*\setminus\CAL E$ and that $(v_j,x_k)\in\CAL E^*\setminus\CAL E$. Thus, $\CAL E^*\setminus\CAL E$ is nonempty.

Furthermore, for any feature node $x_k\in\CAL V_\CAL X$, since any feature edge $(v_i,x_k)\in\CAL E_\CAL X$ ensures $\BM X[v_i,k]=1$, then we have
\AL{
\BM X^*[x_k,k]&=\frac1{|\CAL E_\CAL X\cap(\CAL V\times\{x_k\})|}\sum_{v_i:(v_i,x_k)\in\CAL E_\CAL X}\BM X[v_i,k]\\
&=\frac1{|\CAL E_\CAL X\cap(\CAL V\times\{x_k\})|}\sum_{v_i:(v_i,x_k)\in\CAL E}1\\
&=\frac1{|\CAL E_\CAL X\cap(\CAL V\times\{x_k\})|}\sum_{v_i:(v_i,x_k)\in\CAL E\cap(\CAL V\times\{x_k\})}1\\
&=1
.}

Finally, for any added feature edge $(v_i,x_k)\in\CAL E^*\setminus\CAL E=\CAL E_\CAL X$,
\AL{
\OP{sim}(v_i,x_k)&=\|\BM X[v_i,:]\land\BM X[x_k,:]\|_\infty\\
&=\max_{k'\in\CAL X}|\BM X[v_i,k']\land\BM X[x_k,k']|\\
&\ge|\BM X[v_i,k]\land\BM X[x_k,k]|\\
&=|1\land1|=1
.}
Since $\OP{hom}(\CAL G)<1$, then
\AL{\OP{sim}(v_i,x_k)\ge1>\OP{hom}(\CAL G).}
Therefore, by Lemma~\ref{lem:add} with
\AL{
\CAL A&:=\{\!\!\{\OP{sim}(v_i,v_j):(v_i,v_j)\in\CAL E\}\!\!\},\\
\CAL B&:=\{\!\!\{\OP{sim}(v_i,x_k):(v_i,x_k)\in\CAL E_\CAL X\}\!\!\},
}
we have
\AL{
\OP{hom}(\CAL G^*)&=\frac1{|\CAL E^*|}\sum_{(u,u')\in\CAL E^*}\OP{sim}(u,u')\\
&=\frac1{|\CAL E\sqcup\CAL E_\CAL X|}\sum_{(u,u')\in\CAL E\sqcup\CAL E_\CAL X}\OP{sim}(u,u')\\
&=\frac{1}{|\CAL A\sqcup\CAL B|}\sum_{z\in\CAL A\sqcup\CAL B}z\\
&>\frac1{|\CAL A|}\sum_{z\in\CAL A}z\\
&=\frac1{|\CAL E|}\sum_{(v_i,v_j)\in\CAL E}\OP{sim}(v_i,v_j)\\
&=\OP{hom}(\CAL G)
.}

\vspace{0.5em}\noindent\textbf{Number of nodes.} Since $|\CAL X|\le O(|\CAL V|)$, then
\AL{|\CAL V_\CAL X|=|\CAL X|\le O(\CAL V).}
It follows that
\AL{|\CAL V^*|&=|\CAL V|+|\CAL V_\CAL X|\\
&\le|\CAL V|+O(|\CAL V|)\\
&=O(|\CAL V|)
.}

\vspace{0.5em}\noindent\textbf{Number of edges.} Since $\BM X$ is a binary matrix, then $\|\BM X\|_1=\|\BM X\|_0\le O(|\CAL E|)$. Hence,
\AL{
|\CAL E_\CAL X|&=\sum_{v_i\in\CAL V}\sum_{x_k\in\CAL V_\CAL X}1_{[(v_i,x_k)\in\CAL E_\CAL X]}\\
&=\sum_{v_i\in\CAL V}\sum_{k\in\CAL X}1_{[(v_i,x_k)\in\CAL E_\CAL X]}\\
&=\sum_{v_i\in\CAL V}\sum_{k\in\CAL X}1_{[\BM X[v_i,k]=1]}\\
&=\sum_{v_i\in\CAL V}\sum_{k\in\CAL X}\BM X[v_i,k]\\
&=\sum_{v_i\in\CAL V}\sum_{k\in\CAL X}|\BM X[v_i,k]|\\
&=\|\BM X\|_1=\|\BM X\|_0\le O(|\CAL E|)
.}
It follows that
\AL{
|\CAL E^*|&=|\CAL E|+|\CAL E_\CAL X|\\
&\le|\CAL E|+O(|\CAL E|)\\
&=O(|\CAL E|)
.}

\end{document}